\documentclass{article}

\usepackage[preprint]{neurips_2025}

\usepackage[utf8]{inputenc} % allow utf-8 input
\usepackage[T1]{fontenc}    % use 8-bit T1 fonts
\usepackage{hyperref}       % hyperlinks
\usepackage{url}            % simple URL typesetting
\usepackage{booktabs}       % professional-quality tables
\usepackage{amsfonts}       % blackboard math symbols
\usepackage{nicefrac}       % compact symbols for 1/2, etc.
\usepackage{microtype}      % microtypography
\usepackage{xcolor}         % colors
\usepackage{multirow}
\usepackage[pdftex]{graphicx}
\usepackage{mathrsfs}
\usepackage{amsmath}
\usepackage{amssymb}
\usepackage{algorithm}
\usepackage{algorithmic}
\usepackage{array}
\usepackage{bm}
\usepackage{caption3}
\usepackage{url}
\usepackage{pifont} 
\usepackage{subfigure}
\usepackage{wrapfig}
\usepackage{makecell}
\usepackage{colortbl}
\usepackage{enumitem}
\usepackage{multirow}

\title{Collaborating Vision, Depth, and Thermal Signals for Multi-Modal Tracking: Dataset and Algorithm}

\author{
 Xue-Feng~Zhu$^1$, Tianyang Xu$^{1,}$\thanks{Corresponding author: \texttt{tianyang.xu@jiangnan.edu.cn}} , Yifan Pan$^1$, Jinjie Gu$^1$, \\
 \textbf{Xi Li}$^2$, \textbf{Jiwen Lu}$^3$, \textbf{Xiao-Jun Wu}$^1$, \textbf{Josef Kittler}$^4$\\
 $^1$ Jiangnan University; $^2$ Zhejiang Univeristy; $^3$ Tsinghua University; $^4$ University of Surrey\\
 }
 
\begin{document}

\maketitle

\begin{abstract}
Existing multi-modal object tracking approaches primarily focus on dual-modal paradigms, such as RGB-Depth or RGB-Thermal, yet remain challenged in complex scenarios due to limited input modalities. 
To address this gap, this work introduces a novel multi-modal tracking task that leverages three complementary modalities, including visible RGB, Depth (D), and Thermal Infrared (TIR), aiming to enhance robustness in complex scenarios. 
To support this task, we construct a new multi-modal tracking dataset, coined RGBDT500, which consists of 500 videos with synchronised frames across the three modalities. 
Each frame provides spatially aligned RGB, depth, and thermal infrared images with precise object bounding box annotations.
Furthermore, we propose a novel multi-modal tracker, dubbed RDTTrack.
RDTTrack integrates tri-modal information for robust tracking by leveraging a pretrained RGB-only tracking model and prompt learning techniques.
In specific, RDTTrack fuses thermal infrared and depth modalities under a proposed orthogonal projection constraint, then integrates them with RGB signals as prompts for the pre-trained foundation tracking model, effectively harmonising tri-modal complementary cues.
The experimental results demonstrate the effectiveness and advantages of the proposed method, showing significant improvements over existing dual-modal approaches in terms of tracking accuracy and robustness in complex scenarios. 
The dataset and source code are publicly available at \url{https://xuefeng-zhu5.github.io/RGBDT500}.
\end{abstract}

\section{Introduction}
% Para.1. Visual object Tracking --> dual-modal object tracking
% Para.2. Current dual-modal object tracking, dataset issues: 
%   (1) object imaging failure in two modality; --> datasets: scene complexity limited
%   (2) Dual-modality tracking methods--dual-modal fusion;limitation in robustness and generalization performance
% Figure.1 --> current RGB-T/RGB-D dataset: two-modal failure scenarios; our dataset. 
% Para.3. explain our dataset and the advantages
% Para.4. explain the tri-modal algorithm

% \begin{figure}
%   \centering
%    \includegraphics[width=1.0\linewidth]{images/motivation.pdf}
%    \caption{(a) illustrates the development of multi-modal tracking datasets, including RGB, RGB-T, RGB-D, and RGB-D-T datasets, with our method (RGBDT500) being the first spatiotemporally aligned RGB-D-T dataset. Fig. 1 (b) presents our tri-modal baseline tracking pipeline, which integrates RGB, Depth, and Thermal modalities, enabling more effective handling of the challenges in multi-modal tracking.}
%    \label{fig:one}
   
% \end{figure}

% \begin{figure*}
%   \centering
%   % \fbox{\rule{0pt}{2in} \rule{0.9\linewidth}{0pt}}
%    \includegraphics[width=1.0\linewidth]{images/xuliezhanshi.pdf}
%    \caption{Partial sequences from the RGBDT500 dataset are shown, with the tracking target highlighted by green bounding boxes.}
%    \label{fig:two}
% \end{figure*}

Visual object tracking aims to automatically localise an object of interest within a video, based on the initially specified object position and scale~\citep{javed2023visual, liu2024spatial, kristan2023first}. 
It is a fundamental research topic in the field of artificial intelligence and computer vision.
Through decades of research, visual object tracking has witnessed substantial progress, driven by continuous advancements in benchmark datasets~\citep{huang2019got, fan2019lasot, zhu2024unimod1k} and tracking algorithms~\citep{xu2021adaptive, xu2023learning, zhu2022robust, xu2020accelerated}.
However, conventional tracking methods mainly rely on visible RGB images, often suffer from reduced effectiveness and robustness under conditions of adverse visibility~\citep{tang2025revisiting}.

To cope with these challenges, recent research has explored the integration of an additional sensing modality to enhance tracking performance in complex and visually degraded scenarios. 
For instance, multi-modal tracking methods, such as RGB-D and RGB-T tracking, leverage complementary information from depth and thermal data, respectively, to improve robustness and accuracy in adverse conditions.
In particular, RGB-D tracking~\citep{zhu2024self,zhu2023rgbd1k} combines an RGB image with a depth map to provide additional spatial and structural cues, enabling more accurate localisation in challenging scenarios such as occlusion, background clutter, and scale variation. 
Similarly, RGB-T tracking~\citep{tang2024generative, xiao2022attribute} integrates RGB and thermal infrared images, compensating for the limitations of visible light in low-illumination scenarios. 

\begin{figure*}[t]
  \centering
   \includegraphics[trim={0mm 115mm 5mm 0mm},clip,width=0.95\linewidth]{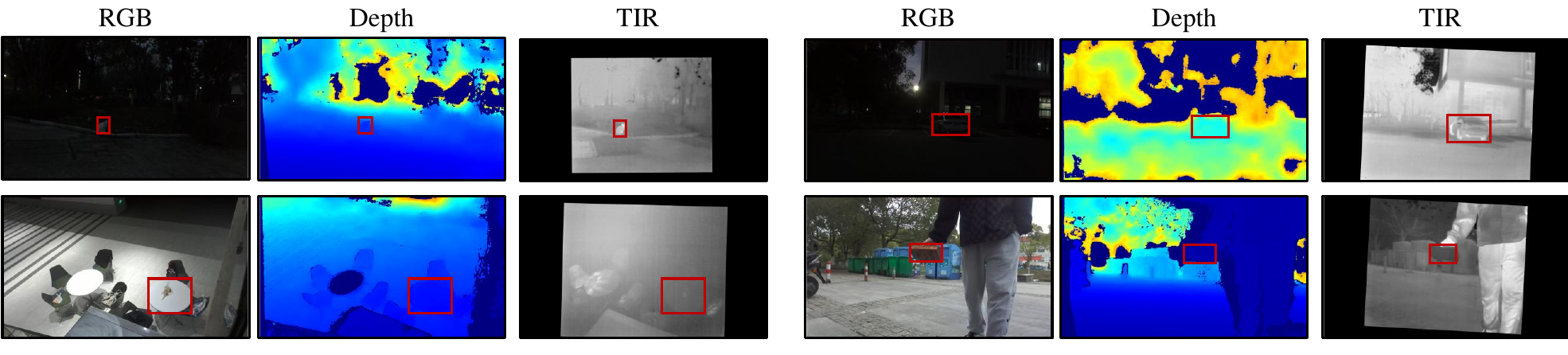}
   \caption{Representative tri-modal samples of RGBDT500. In each sample, at least one modality is affected by specific challenges, highlighting the need for more general multi-modal fusion.}
   \label{fig:rep_samples}
\end{figure*}

Despite these advantages, dual-modal RGB-D and RGB-T trackers still struggle to perform reliably in real-world environments characterised by simultaneous challenges such as poor lighting, occlusions, adverse weather, etc.
As shown in Fig.~\ref{fig:rep_samples}, in some scenarios, both dual-modal RGB-T and RGB-D tracking frameworks may experience modality-specific information degradation due to factors such as thermal crossover, sensor noise, or depth ambiguity in complex scenes.
To address these limitations, we construct RGBDT500, a comprehensive tri-modal tracking dataset comprising synchronised RGB, depth, and thermal infrared video sequences.

Specifically, RGBDT500 includes a total of 500 video sequences, with 400 used for training and 100 reserved for testing.
Each frame in the dataset provides spatially aligned RGB, depth, and thermal infrared images.
% On the one hand, the training set of RGBDT500 consists of {\textcolor{red}{xxx}} RGB-Depth-Thermal (RGB-D-T) image triplets.
% To minimize annotation costs and satisfy training requirement, we employ K-means clustering~\citep{hartigan1979algorithm} to select the most representative frames from each training sequence, which are subsequently annotated with object bounding boxes.
% On the other hand, the test set contains 43,650 RGB-D-T image triplets in total,  each annotated with an object bounding box.
The training set of RGBDT500 consists of about 160K RGB-Depth-Thermal (RGB-D-T) image triplets, while the test set contains 43.7K RGB-D-T image triplets, each annotated with an object bounding box.
Compared to existing dual-modal RGB-D and RGB-T tracking datasets, the three modalities of RGBDT500 offer richer information for object localisation. 
Based on the developed dataset, we naturally extend to a new multi-modal tracking task, tri-modal tracking, which requires leveraging three complementary modalities, including RGB, depth and thermal infrared, for robust tracking.
By effectively integrating RGB, depth, and thermal infrared cues, tri-modal tracking enables adaptive enhanced performance in complex scenarios.

Although tri-modal tracking data provides more advantageous information, it also introduces additional challenges for current multi-modal tracking paradigms.
Existing multi-modal tracking approaches primarily focus on fusing RGB with a single additional modality, making them inadequate for directly handling tri-modal data. 
This limitation is further compounded by the substantial differences among the three modalities. 
For tracking, RGB provides rich texture and colour details, thermal infrared highlights salient heat-emitting objects, while depth captures spatial and geometric structure.
To address these challenges and validate the effectiveness of the RGBDT500 dataset, we propose a straightforward and effective baseline tracker, named RDTTrack.

In detail, to leverage the representation power of the pre-trained RGB tracking model and effectively integrate tri-modal information for object localisation, RDTTrack incorporates a prompt learning mechanism.
Specifically, to fuse depth and thermal infrared cues, it utilises a feature projection to enforce orthogonality between depth and thermal infrared features.
This orthogonal projection aims to reduce feature redundancy and enhance the reliability of the fused representation.
Subsequently, fused depth and thermal features are integrated with RGB features as prompts for the pre-trained OSTrack~\citep{ye2022joint} model, effectively leveraging tri-modal complementary cues for robust tracking.
For RDTTrack training, the pre-trained OSTrack model is kept frozen, and only the prompt learning module is fine-tuned using the 400 training sequences from the RGBDT500 dataset. 
Extensive experiments, including ablation studies and comparative evaluations, demonstrate the effectiveness and the competitive performance of the proposed baseline tracker.

In summary, our contributions are as follows:
\begin{itemize}
\item We introduce a novel multi-modal object tracking task that incorporates three modalities to ensure robust tracking performance, thereby extending existing dual-modal tracking and promoting further progress in the field.
\item  We present RGBDT500, a dataset with temporally and spatially aligned RGB, depth, and thermal modalities, establishing a data foundation for advancing tri-modal tracking research.
\item We propose a novel multi-modal baseline tracker, RDTTrack, which integrates prompt learning with orthogonality constraints to enable effective fusion of tri-modal information.
\item Extensive experiments on the proposed RGBDT500 benchmark demonstrate the generalisation and effectiveness of our baseline RDTTrack.
\end{itemize}

\section{Related Work}\label{relatedwork}
Recent visual tracking research has evolved from uni-modal to multi-modal approaches, enabling more robust performance in complex environments. 
This section provides an overview of research progress in multi-modal tracking, specifically focusing on relevant datasets, RGB-D and RGB-T tracking methodologies.

\subsection{Multi-Modal Tracking Datasets}

Developing multi-modal tracking datasets is essential for both advancing and rigorously evaluating tracking algorithms. 
To date, most existing datasets have concentrated on RGB-D and RGB-T modalities.
Among RGB-D tracking datasets, DepthTrack~\citep{yan2021depthtrack} stands out as a significant benchmark, comprising 200 video sequences. 
More recently, the RGBD1K dataset~\cite{zhu2023rgbd1k} has been introduced, featuring 1,050 sequences, with object bounding box annotations provided for around 720K RGB-D frames. 
These large-scale datasets have been instrumental in advancing the development of RGB-D tracking approaches. 
Additionally, other important benchmarks, such as CDTB~\cite{lukezic2019cdtb}, PTB\cite{song2013tracking}, and STC~\cite{xiao2017robust}, are also valuable for evaluating RGB-D tracking algorithms.

In RGB-T tracking, several influential datasets have been introduced, including GTOT~\cite{li2016learning}, RGBT234~\cite{li2019rgb}, and LasHeR~\cite{li2021lasher}. 
GTOT serves as the first benchmark dataset for RGB-T tracking, comprising 50 RGB-T video sequences. 
Building upon the GTOT benchmark, RGBT234 extends the scale to 234 RGB-T sequences, providing broader coverage of diverse tracking scenarios.
To date, LasHeR represents the most extensive RGB-T tracking dataset, containing 1,244 RGB-T sequences and over 730K paired frames.
The thermal modality, which is particularly effective under low-light and adverse illumination conditions, enhances tracking robustness in real-world environments. 
The release of these datasets has significantly contributed to the advancement of RGB-T tracking.

Despite these advancements, most existing multi-modal datasets are restricted to RGB-D or RGB-T modality combinations, lacking support for more general multi-modal tracking.  
Additionally, ensuring accurate temporal and spatial alignment between modalities and adequately covering the broad range of object categories across diverse tracking conditions remains a significant challenge.
To address these gaps, we introduce RGBDT500, the first dataset featuring spatiotemporally aligned RGB, depth, and thermal modalities for visual tracking.
% RGBDT500 contains 500 high-quality video sequences, spanning a broad spetrum of object categories and complex scenarios.
% In short, our RGBDT500 offers a robust foundation for the development and evaluation of tri-modal tracking approaches.

\subsection{Multi-Modal Tracking Methodologies}
RGB-D tracking leverages complementary information from both RGB and depth modalities, enabling more comprehensive scene understanding for tracking. 
A variety of trackers have been proposed to realise effective fusion and interaction between RGB and depth modalities.
For example, DAL~\citep{qian2021dal} integrates depth information into RGB features based on the correlation filter-based tracking framework.
% TSDM~\citep{zhao2021tsdm} introduces dedicated depth modules to mitigate background interference. 
Recent trackers, such as DeT~\citep{yan2021depthtrack}, adapt RGB-only tracking architectures~\citep{danelljan2019atom, bhat2019learning} by fusing feature maps through pixel-wise operations.
SPT~\citep{zhu2023rgbd1k} utilises separated transformer encoders for RGB and depth, followed by a dedicated fusion module.
ARKitTrack~\citep{zhao2023arkittrack} further advances fusion by encoding depth into bird’s eye view representations and performing cross-view fusion.
Collectively, these approaches represent ongoing progress in RGB-D tracking, aiming to enhance multi-modal feature extraction, fusion, and adaptation for improved tracking robustness.

RGB-T tracking leverages the complementary strengths of RGB and thermal modalities to enhance tracking performance under challenging conditions such as low-light, nighttime environments. 
Early efforts primarily focus on convolutional neural network-based fusion strategies~\citep{tang2025omni,he2016deep}.
For example, mfDiMP~\citep{zhang2019multi} introduces an end-to-end framework for RGB-T fusion, while MANet~\citep{li2019multi} employes a three-way adapter network to extract modality-specific and shared features between two modalities.
% In addition, to better exploit the , adaptive fusion mechanisms have been proposed. 
% CAT~\citep{li2020challenge} employed a multi-branch structure to separately process RGB and thermal modality information, then fused them through an interaction mechanism to address issues of modality sharing and specificity. 
APFNet~\citep{xiao2022attribute} introduces an adaptive feature fusion mechanism that fine-tunes attributes between different modalities, thus enhancing the robustness and accuracy of the algorithm.
Furthermore, with the advent of Vision Transformers~\citep{dosovitskiy2020image}, RGB-T tracking technology has seen further innovations. 
For example, by introducing visual prompt learning~\cite{zhu2023visual,wu2024single,wang2024temporal,hong2024onetracker} into the Transformer-based tracking frameworks, auxiliary thermal information is fused into pre-trained RGB-only models, significantly improving tracking performance. 
Additionally, BAT~\cite{cao2024bi} and TBSI~\cite{hui2023bridging} propose feature bridging mechanisms to strengthen cross-modal interactions.

It is also noteworthy that research on pixel-level RGB-T image fusion provides valuable guidance for cross-modal information integration in tracking~\citep{liu2024promptfusion, cheng2025one}. 
The comprehensive review by \citep{liu2024infrared}, which systematically explores the field from data compatibility to task adaptation, offers a solid theoretical foundation for understanding how to effectively fuse pixel-level heterogeneous modalities to serve high-level vision tasks~\citep{jiang2024multispectral, liu2022attention}. 
For instance, CoCoNet~\citep{liu2024coconet} introduces a multi-level feature ensemble for multi-modal image fusion, significantly advancing the performance in both image fusion and downstream object detection tasks.
Through pixel-level fusion, complementary information from RGB and TIR modalities can be effectively aligned and enhanced~\citep{song2025refinefuse}, thereby providing a more robust and information-rich joint representation for subsequent trackers.

However, most current multi-modal tracking methods are predominantly designed for dual-modality scenarios, typically combining RGB with either depth or thermal information.
As such, they are not well-suited to directly process and integrate tri-modal data due to architectural limitations and the distinct characteristics of each modality.
In contrast, our proposed RDTTrack baseline is capable of jointly leveraging RGB, depth, and thermal inputs, thus improving tracking robustness. 
In addition, its architecture is tailored to handle the heterogeneity across modalities.
% Consequently, it demonstrates superior adaptability and robust performance when evaluated on our newly proposed RGBDT500 dataset, compared to prior RGB-D or RGB-T trackers.

\begin{figure*}
  \centering
   \includegraphics[width=1.0\linewidth]{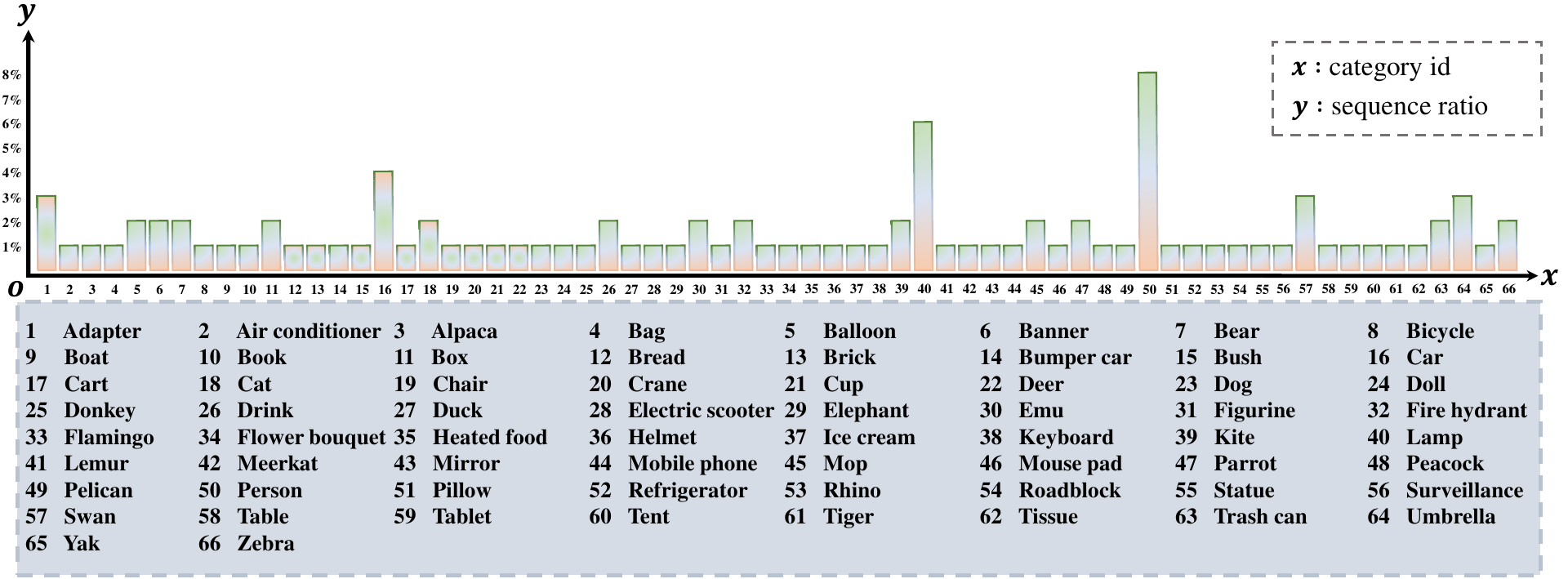}
   \caption{The object category distribution of the RGBDT500 test set.}
   \label{fig:classes}
\end{figure*}

\begin{table*}[t]
\footnotesize
\centering
\caption{A comparison of RGBDT500 with related RGB-only, RGB-T, and RGB-D tracking datasets.}
\resizebox{0.95\linewidth}{!}{
\begin{tabular}{ccccccccc}
\toprule
\multicolumn{1}{c}{\multirow{2}{*}{\textbf{Dataset}}}  & \multicolumn{3}{c}{\textbf{Modalities}}  & \multicolumn{1}{c}{\multirow{2}{*}{\textbf{Sequences}}} &  \multicolumn{1}{c}{\multirow{2}{*}{\textbf{Frames}}}  & \multicolumn{1}{c}{\multirow{2}{*}{\textbf{Classes}}} & \multicolumn{1}{c}{\multirow{2}{*}{\textbf{Training Set}}} & \multicolumn{1}{c}{\multirow{2}{*}{\textbf{Publication}}} \\ 
% \cline{2-4}
 & \textbf{RGB} & \textbf{Depth} & \textbf{TIR} & & & & & \\
\midrule
UAV123~\cite{benchmark2016benchmark} &\ding{52} & \ding{56} &\ding{56} & 123 & 113K & - & \ding{56} & ECCV 2016 \\
GOT-10k~\cite{huang2019got} &\ding{52} & \ding{56} &\ding{56} & 10K & 1.5M & 563 & \ding{52} & IEEE TPAMI 2019 \\
LaSOT~\cite{fan2019lasot} & \ding{52} & \ding{56} &\ding{56} & 1400 & 3.5M & 70 & \ding{52} & CVPR 2019 \\
\midrule
GTOT~\cite{li2016learning} & \ding{52} & \ding{56} &\ding{52} & 50 & 15.8K & 9 & \ding{56} & IEEE TIP 2016 \\
RGBT210~\cite{li2017weighted} & \ding{52} & \ding{56} &\ding{52} & 210 & 104.7K & 22 & \ding{56} & ACM MM 2017 \\
RGBT234~\cite{li2019rgb} & \ding{52} & \ding{56} &\ding{52} & 234 & 116.7K & 22 & \ding{56} & PR 2019 \\
LasHeR~\cite{li2021lasher}  & \ding{52} & \ding{56} &\ding{52} & 1224 & 734.8K & 32 & \ding{52} & IEEE TIP 2021 \\
VTUAV~\cite{zhang2022visible}  & \ding{52} & \ding{56} &\ding{52} & 500 & 1.7M & 13 & \ding{52} & CVPR 2022 \\
\midrule
PTB~\cite{song2013tracking} & \ding{52} & \ding{52} &\ding{56} & 100 & 21K & 26 & \ding{56} & CVPR 2013\\
CDTB~\cite{lukezic2019cdtb} & \ding{52} & \ding{52} &\ding{56} & 80 & 101.9K & 21 & \ding{56} & ICCV 2019\\
DepthTrack~\cite{yan2021depthtrack} & \ding{52} & \ding{52} &\ding{56} & 200 & 294.5K & 90 & \ding{52} & ICCV 2021\\
ARKitTrack~\cite{zhao2023arkittrack} & \ding{52} & \ding{52} &\ding{56} & 455 & 229.7K & 144 & \ding{52} & CVPR 2023\\
\midrule
\rowcolor{gray!20}RGBDT500 & \ding{52} & \ding{52} &\ding{52} & 500 & 203.7K & 66 & \ding{52} & NeurIPS 2025\\
\bottomrule
\end{tabular}}
\label{table:table1}
\end{table*}

\section{The RGBDT500 Dataset}
To promote the development of more general multi-modal object tracking and to support tri-modal tracking task, this work introduces the RGBDT500 dataset.
The RGBDT500 contains three modalities, including RGB, depth and thermal infrared, encompassing a wide range of object categories and challenging scenarios for tracking.

\subsection{Dataset Details}
The RGBDT500 dataset comprises 400 training sequences, and 100 test sequences, with approximately 160K tri-modal frames in the training set and 43.7K tri-modal frames in the test set.
For each tri-modal frame, the RGB and depth images are captured using the ZED stereo camera, while the thermal infrared image is recorded using a separate LGCS121 thermal camera.
The ZED stereo camera delivers time-synchronised and pixel-aligned RGB and depth image pairs.
To resolve the resolution mismatch between the thermal infrared images and the RGB/depth modalities, we employ a manually feature point matrix mapping approach that aligns the target region and surrounding pixels across all three modalities.
Furthermore, all tri-modal images are stored in PNG format with a uniform resolution of $1920\times1080$.

The RGBDT500 dataset includes a diverse range of object categories, covering over 66 classes, including household items, animals, and vehicles, as shown in Fig.~\ref{fig:classes}. 
In constructing the categories, we strategically consider the unique strengths and limitations of each modality. 
For example, sequences featuring objects in low-light conditions or containing high-temperature objects are deliberately captured to emphasise the advantages of the thermal modality. 
Conversely, we capture several video sequences featuring planar objects with minimal geometric depth variation to intentionally limit the contribution of the depth modality, thereby encouraging greater reliance on RGB and thermal infrared cues.
To further enhance the dataset’s complexity and realism, we capture some sequences exhibiting modal perturbations, such as inconsistent viewpoints among RGB, depth, and thermal infrared streams.
Collectively, these design choices make RGBDT500 a challenging and comprehensive benchmark for evaluating the robustness of tri-modal tracking algorithms.
In addition, to preserve privacy, we have applied EgoBlur~\cite{raina2023egoblur} model to  blur all recognizable faces and license plates present in the images.

Table~\ref{table:table1} presents a comparison between the proposed RGBDT500 dataset and related visual tracking datasets.
As shown, compared to the current multi-modal tracking dataset, the RGBDT500 dataset offers several key advantages.
It provides spatially and temporally aligned RGB, depth and TIR modalities for tri-modal tracking, which existing datasets limited to RGB-D and RGB-T pairs cannot support.
Besides, compared to most RGB-D and RGB-T tracking datasets, RGBDT500 offers a clear advantage in the number of sequences, total frames and object classes.

\subsection{Data Annotation}
The RGBDT500 is annotated with accurate object bounding boxes for frames containing the object of interest, providing reliable ground truth for developing and assessing tracking algorithms.
For the training set, to minimise annotation costs and while meeting requirements for training, we employ K-means clustering~\citep{hartigan1979algorithm} to select the most representative frames from each training sequence, which are subsequently annotated with object bounding boxes.
Besides, the test set includes 100 sequences, encompassing 43.7K RGB-D-T image triplets with dense bounding box annotations, thereby facilitating rigorous and detailed performance assessment.
For the bounding box annotation, we adopt the top-left corner coordinates $(x, y)$, along with the width $w$ and height $h$ of the target's bounding box, to represent the ground truth in the format $[x, y, w, h]$.

\subsection{Evaluation Metrics}
For evaluation, we follow the One-Pass Evaluation (OPE) protocol~\citep{wu2015object} to assess tracking performance on RGBDT500. Specifically, we compute the Distance Precision (DP) and the Area Under the Curve (AUC) of the success plot to quantitatively measure the effectiveness of multi-modal tracking methods.
Specifically, the precision is measured by computing the distance between the predicted and the ground-truth bounding boxes.
By varying a predefined distance threshold to determine successful tracking, a precision plot can be generated to illustrate tracker performance.
Based on the precision plot, DP is defined as the percentage of frames in which the prediction error falls within a threshold of 20 pixels.
In addition, tracking success can also be evaluated by determining whether the overlap ratio between the predicted and ground-truth bounding boxes exceeds a predefined threshold.
The success plot illustrates the proportion of successful frames across a range of overlap thresholds from 0 to 1. 
Then, the AUC of the success plot is computed to quantitatively assess trackers.

%  figure: pipeline
\begin{figure*}[t]
  \centering
   \includegraphics[width=0.95\linewidth]{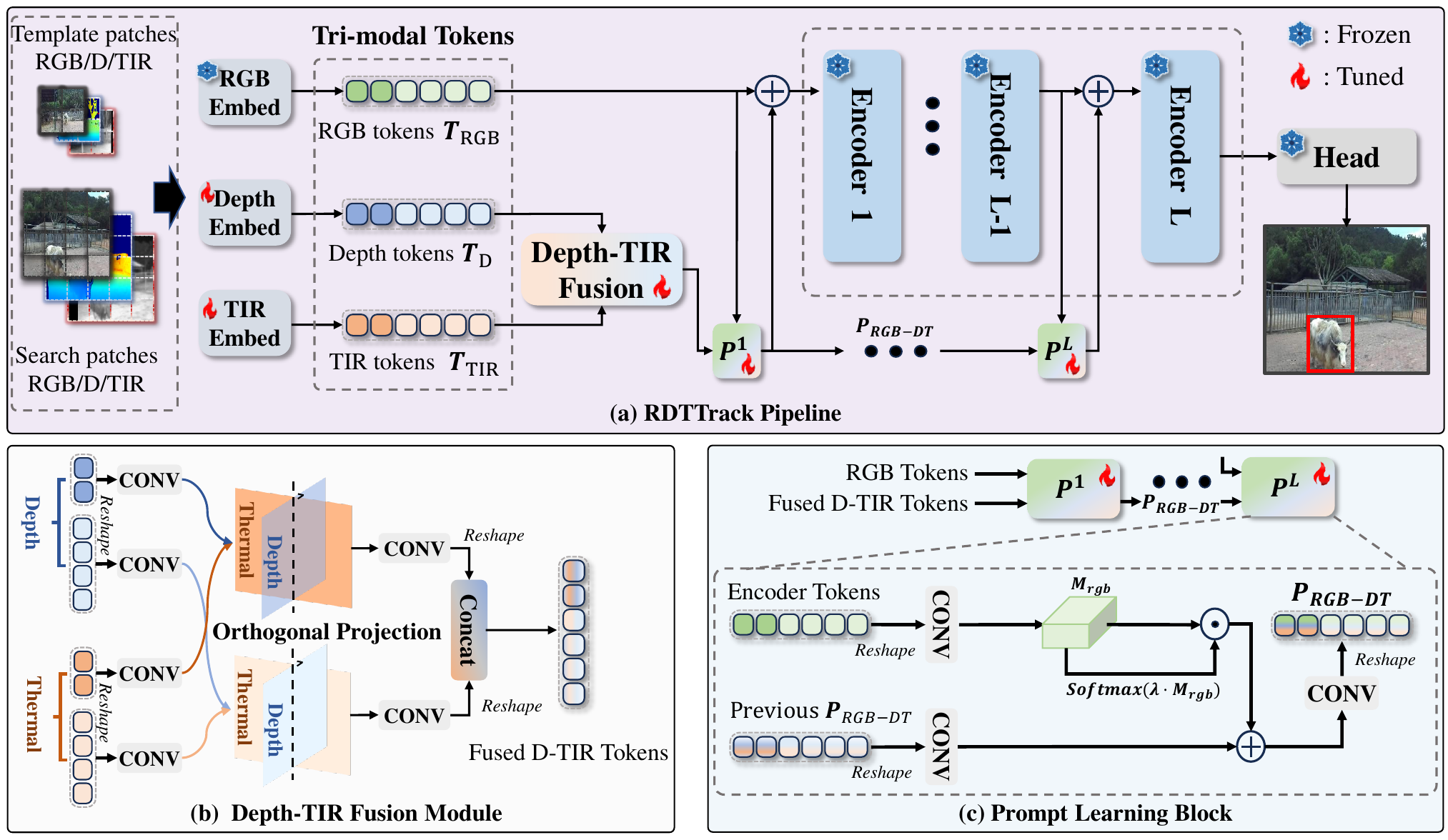}
   \caption{An overview of the pipeline and architecture of RDTTrack. (a) illustrates the overall tracking pipeline; (b) depicts the detailed structure of the Depth-TIR fusion module; and (c) presents the architecture of the prompt learning block.}
   \label{fig:pipeline}
\end{figure*}

\section{The Tri-Modal Tracking Baseline}

To support tri-modal object tracking on RGBDT500 and promote progress toward more general multi-modal tracking, we design a specialised tracking algorithm, RDTTrack, capable of effectively integrating RGB, depth, and thermal infrared information.
The RDTTrack is developed by extending a pre-trained RGB-only tracker, OSTrack~\cite{ye2022joint}, with a specially designed prompt learning module that enables integration of tri-modal cues for tracking.

\subsection{Overall Architecture}
The overall pipeline and architecture of RDTTrack are illustrated in Fig.~\ref{fig:pipeline}.
The pre-trained OSTrack model consists of a standard patch embedding module, $L$ Transformer encoder layers and a bounding box prediction head.
As shown in Fig.~\ref{fig:pipeline}(a), firstly, the tri-modal template patches ($\bm{Z}_M \in \mathbb{R}^{3 \times H_z \times W_z}$, where $M\in\{\mathrm{RGB, D, TIR}\}$) and search patches ($\bm{X}_M\in \mathbb{R}^{3 \times H_x \times W_x}$) are first processed through a patch embedding layer with positional encoding to produce RGB, depth and TIR template tokens $\bm{T}_M^Z \in \mathbb{R}^{C \times h_Zh_Z}$ and search tokens $\bm{T}_M^X \in \mathbb{R}^{C \times h_Xh_X}$ respectively.
It is worth noting that the single-channel depth and TIR images are converted into three-channel representations similar to RGB images.
Then the corresponding template and search tokens for each modality are concatenated to construct the input tokens $\bm{T}_M \in \mathbb{R}^{C \times (h_Zh_Z+h_Xh_X)}$ of each modality, calculated as:
\begin{small}
\begin{equation}
\mathbf{T}_{M} = [(\mathrm{PE}(\bm{Z}_M)+ \mathrm{Pos_Z}) \underset{2}{\Vert}  (\mathrm{PE}(\bm{X}_M)+ \mathrm{Pos_X})], M\in\{\mathrm{RGB, D, TIR}\},
\label{eq1}
\end{equation}
\end{small}
where $\mathrm{PE}$ is the patch embedding operation, and $\mathrm{Pos_Z}$ and $\mathrm{Pos_X}$ are positional encodings. [·$\underset{2}{\Vert}$·] denotes the concatenation operator along the token dimension (the second dimension).

Subsequently, the depth and thermal infrared tokens are integrated as auxiliary multi-modal information to enhance the pre-trained RGB-only tracking OSTrack model.
Specifically, the RGB tokens, together with the fused depth and thermal infrared (D-TIR) tokens, are fed into a prompt learning block designed to learn effective multi-modal visual prompts.
Then, the multi-modal visual prompts are added with the RGB tokens and input to the $L$-layer pre-trained vision Transformer encoder for feature extraction and interaction.
The learned multi-modal visual prompts are subsequently added to the RGB tokens and fed into the $L$-layer pre-trained vision Transformer encoder for feature extraction and interaction. 
Each encoder layer comprises a multi-head self-attention mechanism, layer normalisation, a feed-forward network, and skip connections. 
At each encoder layer, the input is constructed by integrating the tokens from the prompt module with the RGB tokens, thereby enabling robust and comprehensive tri-modal feature interaction.
The output of the final encoder layer is passed to the prediction head, which produces the tracking results.

For the training of RDTTrack, we freeze all parameters of the RGB streams, and fine-tune only the Depth-TIR fusion module and all prompt learning blocks.
The overall loss function of RDTTrack is a combination of focal loss~\cite{lin2017focal} for classification,  $L_{1}$ loss and the GIoU loss~\cite{rezatofighi2019generalized} for localisation, calculated as: $\mathcal{L}= L_{\text{CLS}} + \lambda_{\text{GIoU}} \cdot L_{\text{GIoU}} + \lambda_{L_{1}} \cdot L_{1}$, where $\lambda_{\text{GIoU}}$ and $\lambda_{L_{1}}$ are two constants.

\subsection{Depth-TIR Fusion Module}

To fully exploit the tracking potential of the auxiliary depth and thermal infrared modalities, we propose a depth-TIR fusion module based on orthogonal projection constraints, as shown in Fig.~\ref{fig:pipeline}(b).
In particular, the depth tokens and TIR tokens are divided into template and search tokens.
% , $\bm{T}_D^Z \in \mathbb{R}^{C \times h_Zh_Z}$, $\bm{T}_D^X \in \mathbb{R}^{C \times h_Xh_X}$, $\bm{T}_{TIR}^Z \in \mathbb{R}^{C \times h_Zh_Z}$, $\bm{T}_{TIR}^X \in \mathbb{R}^{C \times h_Xh_X}$. 
These tokens are then reshaped into 2D spatial feature maps for further processing.
To obtain compact and discriminative representations for each modality, two $1\times1$ convolutional layers are independently applied to the depth and TIR inputs.
Then, depth and TIR inputs are processed through an orthogonal projection, calculated as:

\begin{small}
\begin{equation}\left\{
\begin{aligned}
&\bm{F}_D^Z = \bm{F}_D^Z - \alpha\cdot\frac{(\bm{F}_D^Z \cdot \bm{F}_{TIR}^Z)}{\|\bm{F}_{TIR}^Z\| + \epsilon} \cdot \bm{F}_{TIR}^Z, \quad
&\bm{F}_{TIR}^Z = \bm{F}_{TIR}^Z - \beta\cdot\frac{(\bm{F}_{TIR}^Z \cdot \bm{F}_{D}^Z)}{\|\bm{F}_{D}^Z\| + \epsilon} \cdot \bm{F}_{D}^Z,\\
&\bm{F}_D^X = \bm{F}_D^X - \alpha\cdot\frac{(\bm{F}_D^X \cdot \bm{F}_{TIR}^X)}{\|\bm{F}_{TIR}^X\| + \epsilon} \cdot \bm{F}_{TIR}^X, \quad
&\bm{F}_{TIR}^X = \bm{F}_{TIR}^X - \beta\cdot\frac{(\bm{F}_{TIR}^X \cdot \bm{F}_{D}^X)}{\|\bm{F}_{D}^X\| + \epsilon} \cdot \bm{F}_{D}^X,\\
\end{aligned}\right.\label{eq2}
\end{equation}
\end{small}
where the $(., .)$ is inner product operation. $\bm{F}_D^Z \in \mathbb{R}^{C \times h_Z \times h_Z}$, $\bm{F}_D^X \in \mathbb{R}^{C \times h_X \times h_X}$, $\bm{F}_{TIR}^Z \in \mathbb{R}^{C \times h_Z \times h_Z}$, $\bm{F}_{TIR}^X \in \mathbb{R}^{C \times h_X \times h_X}$ are the template and search region feature maps of depth and TIR modalities, respectively. $\alpha$ and $\beta$ are two learnable constant number. $\epsilon$ is a very small fixed constant.
Through the above process, the orthogonal features of the depth and TIR modalities are effectively extracted, enabling them to complement each other and provide richer, more discriminative information for robust tracking.

Subsequently, $\bm{F}_D^Z$ and $\bm{F}_{TIR}^Z$, as well as $\bm{F}_D^X$ and $\bm{F}_{TIR}^X$, are concatenated along the channel dimension, respectively.
Afterwards, a 1×1 convolution is applied to each concatenated feature map, which is then reshaped back into tokens.
The resulting dual-modal template tokens and search tokens are finally concatenated along the token sequence dimension to form a unified D-TIR fused representation $\bm{T}_\mathrm{{D-TIR}}\in \mathbb{R}^{C \times (h_Zh_Z+h_Xh_X)}$. 
These operations are computed as:
\begin{small}
\begin{equation}\left\{
\begin{aligned}
&
\bm{F}_{D-TIR}^Z = \mathrm{Conv}([\bm{F}_D^Z \underset{1}{\Vert} \bm{F}_{TIR}^Z]), \quad
\bm{F}_{D-TIR}^X = \mathrm{Conv}([\bm{F}_D^X \underset{1}{\Vert} \bm{F}_{TIR}^X]),\\
&\bm{T}_{D-TIR} = [\mathrm{Reshape}(\bm{F}_{D-TIR}^Z) \underset{2}{\Vert} \mathrm{Reshape}(\bm{F}_{D-TIR}^X)],\\
\end{aligned}\right.\label{eq3}
\end{equation}
\end{small}
where  [·$\underset{1}{\Vert}$·] denotes the concatenation operator along the channel dimension (the first dimension).

\subsection{Multi-Modal Prompt Learning Blocks}
Fig.~\ref{fig:pipeline}(c) presents the architecture of the multi-modal prompt learning module.
It is designed to extract complementary information between the RGB modality and the fused auxiliary D-TIR modalities to produce more informative visual prompts. 
The prompt learning block follows the design proposed in~\citep{zhu2023visual}.
The $l$-th  prompt learning block $\mathrm{Propmt}^l()$ takes as input the tokens $\bm{H}^{l-1}$ from $(l-1)$-th Transformer encoder layer along with the multi-modal prompts $\bm{P}^{l-1}$ produced by the preceding prompt block, as:
\begin{align}
\bm{P}^l = \mathrm{Propmt}^l\left(\bm{H}^{l-1}, \bm{P}^{l-1}\right), \quad l = 1, 2, \dots, L.
\end{align}

In the prompt learning block, $\bm{P}^{l-1}$ and $\bm{H}^{l-1}$ are projected to a space of reduced channel dimension using a $1\times1$ convolutional layer.
Then the embeddings $\bm{H}^{l-1}$ are enhanced by a spatial fovea operation, which adopts a $\lambda$-smoothed softmax across all the spatial dimensions.
Finally, the multi-modal prompt embeddings are generated by adding $\bm{P}^{l-1}$ to the enhanced $\bm{H}^{l-1}$ and a $1\times1$ convolutional layer.
The detailed operations are calculated as:
\begin{small}
\begin{equation}\left\{
\begin{aligned}
&\bm{A}_{\mathrm{RGB}} = \mathrm{Conv}\left(\bm{H}^{l-1}\right), \quad 
\bm{A}_{P} = \mathrm{Conv}\left(\bm{P}^{l-1}\right)
\\
&\bm{A}_\mathrm{RGB}^e = \bm{A}_\mathrm{RGB} \odot \bm{A}_{\mathrm{fovea}}, \quad
\bm{A}_{\mathrm{fovea}} = \left\{ 
\frac{e^{\bm{A}_{\mathrm{RGB}}[i, j]}}{\sum e^{\bm{A}_{\mathrm{RGB}}[i, j]}} \lambda 
\right\}
\\
&\bm{P}^{l} = \mathrm{Conv}\left(\bm{A}_{\mathrm{RGB}}^e+\bm{A}_{P}\right)
\end{aligned}\right.,\label{eq2}
\end{equation}
\end{small}
$\odot$ denotes element-wise multiplication. For the first prompt learning block, its input is the RGB tokens $\bm{T}_{\mathrm{RGB}}$ and the D-TIR fused representation $\bm{T}_{\mathrm{D-TIR}}$.
For more detailed information of the prompt learning block, readers are referred to the reference~\citep{zhu2023visual}.

\section{Experiments}
\subsection{Experimental Settings}
Our proposed RDTTrack is implemented in Python 3.8 with PyTorch 1.12, and all training and evaluation procedures are conducted on a single NVIDIA RTX 3090 GPU. 
The RDTTrack is fine-tuned on the RGBDT500 dataset for 60 epochs, with each epoch containing 60K tri-modal samples.
The frozen parameters of RDTTrack are initialised using the pre-trained weights of the OSTrack model.
The initial learning rate is set to $4e-5$ and is decreased by a factor of 10 after the 48-th epoch.
All evaluations are performed using the standard OPE tracking protocol, including DP, AUC and tracking speed Frames Per Second (FPS) metrics.

\begin{table*}[t]
\footnotesize
\centering
\caption{Comparison of RDTTrack with multiple SOTA trackers on RGBDT500.}
\begin{tabular}{ccccc}
\toprule
Tracker        & Input Modality  & AUC    & DP        & Publication Year \\ \hline
ATOM~\cite{danelljan2019atom}           & RGB               & 0.649   & 0.695      & 2019          \\
DiMP~\cite{bhat2019learning}           & RGB               & 0.662   & 0.710      & 2019          \\
PrDiMP~\cite{danelljan2020probabilistic}         & RGB               & 0.676   & 0.698      & 2020          \\
ToMP~\cite{mayer2022transforming}           & RGB               & 0.710   & 0.747      & 2022          \\ 
OSTrack~\cite{ye2022joint}& RGB                              & 0.694  &  0.736   & 2022  \\
SeqTrack~\cite{chen2023seqtrack}           & RGB               & 0.719   & 0.766      & 2023          \\ 
MixFormer~\cite{cui2024mixformer}           & RGB               & 0.732   & 0.781      & 2024          \\ \midrule
DeT\_ATOM~\cite{yan2021depthtrack}      & RGB+D         & 0.639   & 0.665      & 2021          \\
DeT\_DiMP~\cite{yan2021depthtrack}       & RGB+D         & 0.667   & 0.700      & 2021          \\
SPT~\cite{zhu2023rgbd1k}            & RGB+D         & 0.706   & 0.757      & 2023          \\
ViPT\_RGBD~\cite{zhu2023visual}           & RGB+D         & 0.720   & 0.759      & 2023          \\
SDSTrack\_RGBD~\cite{hou2024sdstrack}& RGB+D        & 0.718   & 0.763      & 2024          \\ UnTrack\_RGBD~\cite{wu2024single}& RGB+D        & 0.733   & 0.776      & 2024          \\\midrule
TBSI~\cite{hui2023bridging}           & RGB+T       & 0.690   & 0.749      & 2023          \\
ViPT\_RGBT~\cite{zhu2023visual}           & RGB+T      & 0.693   & 0.752      & 2023          \\
SDSTrack\_RGBT~\cite{hou2024sdstrack}       & RGB+T       & 0.666   & 0.708      & 2024          \\
BAT~\cite{cao2024bi}            & RGB+T       & 0.713   & 0.782      & 2024          \\ 
UnTrack\_RGBT~\cite{wu2024single}& RGB+T        & 0.732   & 0.790      & 2024          \\
DCEvo+OSTrack~\cite{liu2025dcevo}& RGB+T        & 0.706   & 0.741      & 2025          \\\midrule
\textbf{RDTTrack(Ours)} & RGB+D+T & \textbf{0.752} &\textbf{0.792} & 2025          \\ 
\bottomrule
\end{tabular}
\label{table:comparison}
\end{table*}

\begin{figure}[t]
\centering
\subfigure[]{
\begin{minipage}[t]{0.48\linewidth}
\centering
\includegraphics[trim={0mm 0mm 0mm 0mm},clip,width=1\linewidth]{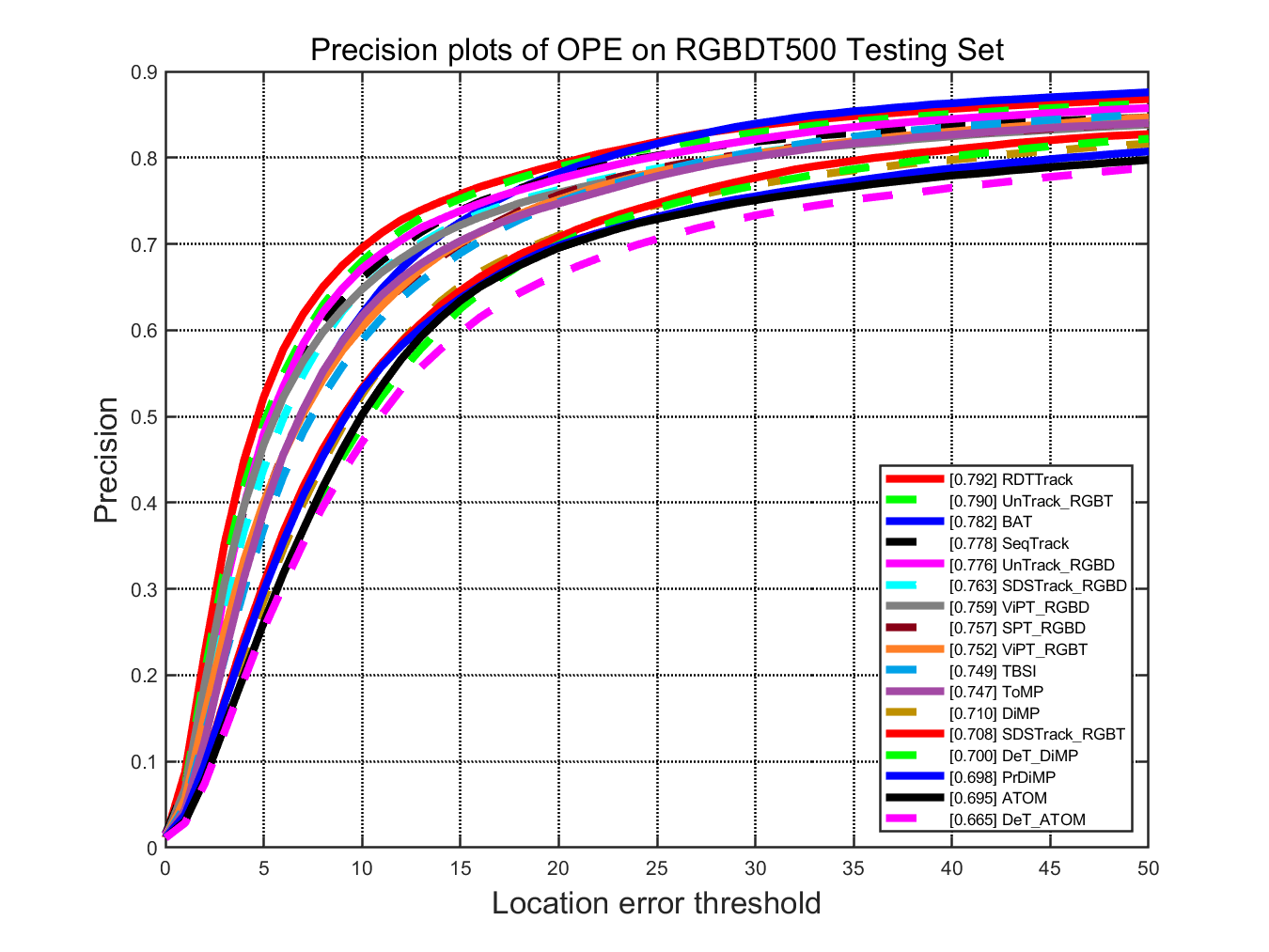}
%\caption{fig1}
\end{minipage}%
}%
\subfigure[]{
\begin{minipage}[t]{0.48\linewidth}
\centering
\includegraphics[trim={0mm 0mm 0mm 0mm},clip,width=1\linewidth]{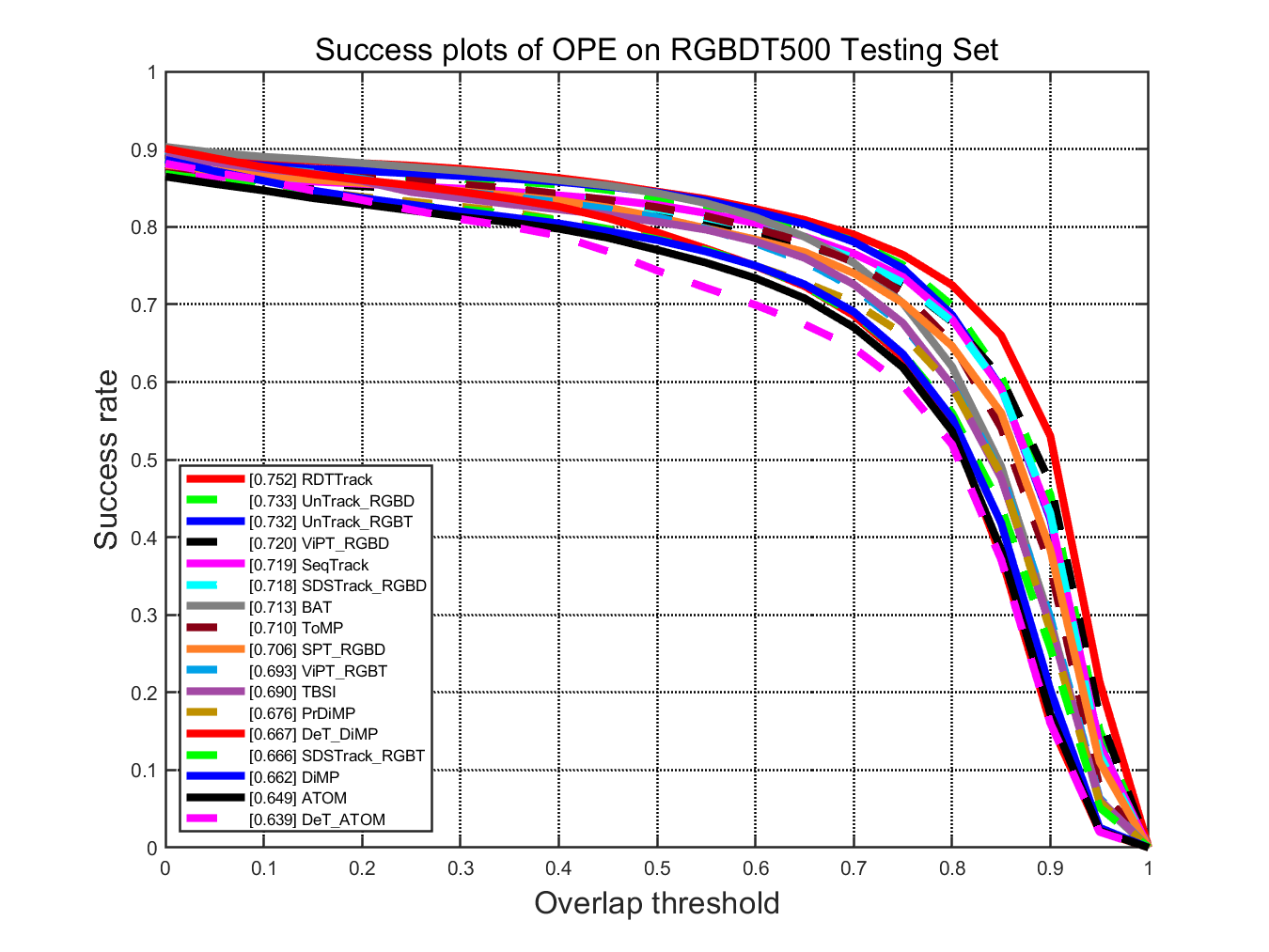}
%\caption{fig2}
\end{minipage}%
}%
\centering
\caption{The precision plots and success plots of trackers on RGBDT500.}
\label{fig:plots}
\end{figure}

\begin{table}[t]
\footnotesize
\centering
\caption{Tracking efficiency comparison of several trackers.}
\begin{tabular}{ccccccc}
\toprule
Tracker & RDTTrack & BAT & TBSI & SDSTrack & Un-Track & ViPT \\\midrule
FPS & \textbf{76.6} & 26.8 & 32.5 & 20.9 & 41.2 & 31.8\\
\bottomrule
\end{tabular}
\label{table:fps}
\end{table}

\subsection{Comparative Experiments}
We evaluate a range of  RGB-only, RGB-D and RGB-T trackers, inclding ATOM~\cite{danelljan2019atom}, DiMP~\cite{bhat2019learning}, PrDiMP~\cite{danelljan2020probabilistic}, ToMP~\cite{mayer2022transforming}, OSTrack~\cite{ye2022joint}, SeqTrack~\cite{chen2023seqtrack}, Mixformer~\cite{cui2024mixformer}, DeT~\cite{yan2021depthtrack}, SPT~\cite{zhu2023rgbd1k}, ViPT~\cite{zhu2023visual}, SDSTrack~\cite{hou2024sdstrack}, TBSI~\cite{hui2023bridging}, BAT~\cite{cao2024bi}, UnTrack~\cite{wu2024single}, and DCEvo~\cite{liu2025dcevo} fusion with OSTrack, and compare them with the proposed RDTTrack.
Table~\ref{table:comparison} reports the tracking performance of various methods on the RGBDT500 test set using AUC and DP metrics. 
Our RDTTrack achieves 0.752 and 0.792 in AUC and DP, respectively.
As shown, the proposed RDTTrack demonstrates superior performance across both metrics, consistently outperforming existing uni-modal and dual-modal trackers.
Specifically, among single-modal RGB-only trackers, the MixFormer achieves the best DP and AUC of 0.732 and 0.781.
By effectively leveraging additional depth and thermal infrared modalities, RDTTrack achieves superior performance compared to MixFormer, with improvements of 2.0\% in AUC and 1.1\% in DP, respectively.

For dual-modal tracking approach, while the best RGB-D and RGB-T trackers outperform single-modal approaches, they remain inferior to our tri-modal tracking framework. 
In detail, ViPT\_RGBD, SDSTrack\_RGBD and UnTrack\_RGBD achieve AUC scores of 0.720, 0.718 and 0.733, and DP scores of 0.759, 0.763 and 0.776, respectively.
In comparison, RDTTrack surpasses these two RGB-D tracking approaches with an AUC improvement of 3.2\%, 3.4\% and 1.9\% respectively, and a DP improvement of 3.3\%, 2.9\% and 1.6\%, respectively.
Among the RGB-T trackers, the UnTrack\_RGBT achieves the optimal AUC of 0.732 and DP of 0.790, outperforming other RGB-T methods like ViPT\_RGBT, SDS\_RGBT and BAT.
In addition, the RGB-T image fusion method DCEvo~\cite{liu2025dcevo} with OSTrack achieves improved performance than RGB-only OSTrack.
It clearly validates the effectiveness of image-level fusion of DCEvo for robust multi-modal tracking.
The proposed tri-modal tracker, RDTTrack, which incorporates an additional depth modality compared to BAT, achieves performance gains of 3.9\% in AUC and 1.0\% in DP, respectively.
In summary, these results validate the effectiveness of RDTTrack’s architecture in fully exploiting the complementary strengths of RGB, depth, and thermal modalities for enhanced tracking performance.

In Fig.~\ref{fig:plots}, we present the precision plots and success plots of trackers on RGBDT500.
The DP and AUC scores are exhibited in the legends of the corresponding plots.
From the curves, it is evident that the proposed RDTTrack consistently outperforms other single-modal and dual-modal trackers, demonstrating superior tracking accuracy and robustness across different evaluation metrics.
Furthermore, we also provide some qualitative results in the supplementary material to intuitively show the advantages of the proposed RDTTrack.

For efficiency comparison, we conduct runtime evaluations of several trackers using FPS as the metric. All experiments are run on an NVIDIA RTX 3090 GPU, and the results are presented in Table~\ref{table:fps}.
As shown, RDTTrack achieves significantly higher runtime efficiency compared to other recent multi-modal trackers, reaching 76.6 FPS. This achievement is owes to the use of lightweight prompt learning and a frozen baseline, which reduces the number of additional parameters and the resulting computational overhead. In addition, the RDTTrack model contains only 0.86M trainable parameters, making it highly efficient for training as well.

\begin{table}[t]
\footnotesize
\centering
\caption{Performance comparison of retrained several trackers.}
\begin{tabular}{cccc}
\toprule
Tracker        & Modality of Training Set & AUC  &DP         \\ \midrule
STARK~\cite{yan2021learning}         & RGB                              & 0.692  &  0.732      \\
SPT\_RGBD~\cite{zhu2023rgbd1k}      & RGB+Depth                        & 0.706  &  0.757      \\
SPT\_RGBT~\cite{zhu2023rgbd1k}      & RGB+Thermal                      & 0.730  &  0.784      \\
\midrule
OSTrack~\cite{ye2022joint}& RGB                              & 0.694  &  0.736     \\
ViPT\_RGBD~\cite{zhu2023visual}     & RGB+Depth                        & 0.719  &  0.762      \\
ViPT\_RGBT~\cite{zhu2023visual}      & RGB+Thermal                      & 0.729  &  0.768      \\ \midrule
RDTTrack & RGB+Depth       & 0.737  & 0.776 \\
RDTTrack & RGB+Thermal     & 0.734   & 0.774  \\
RDTTrack & RGB+Depth+Thermal       & \textbf{0.752} &\textbf{0.792} \\
\bottomrule
\end{tabular}
\label{table:three}
\end{table}

% \begin{table}[t]
% \footnotesize
% \centering
% \caption{Ablation study of modality used in RDTTrack.}
% \begin{tabular}{cccc}
% \toprule
% Tracker  & Input Modality & AUC    &DP        \\ \midrule
% RDTTrack & RGB+Depth         & 0.737  & 0.776       \\
% RDTTrack & RGB+Thermal       & 0.734   & 0.774      \\
% RDTTrack & RGB+Depth+Thermal & \textbf{0.752} &\textbf{0.792} \\
% \bottomrule
% \end{tabular}
% \label{table:four}
% \end{table}

\subsection{Ablation Studies}
In Table~\ref{table:three}, we present the performance of SPT and ViPT on the RGBDT500 test set, following their retraining on the RGBDT500 training set.
With training on dual-modal RGB-D and RGB-T data of RGBDT500, the SPT and ViPT achieve oblivious performance improvements compared to their single-modal baselines STARK~\cite{yan2021learning} and OSTrack~\cite{yan2021learning}.
Nevertheless, our proposed RDTTrack, which integrates RGB, depth, and thermal modalities, consistently outperforms the retrained SPT and ViPT models, demonstrating the advantage of tri-modal fusion.

\begin{wraptable}{r}{6.0cm}
\footnotesize
\centering
\caption{The ablation study on the Depth-TIR fusion module.}
\begin{tabular}{ccc}
\toprule
Tracker                       & AUC &DP            \\ 
\midrule
w/o D-TIR OP & 0.733  &0.773        \\
w/o  $\alpha \& \beta$ & 0.739  &0.779        \\
RDTTrack                         & \textbf{0.752} & \textbf{0.792} \\ 
\bottomrule
\end{tabular}
\label{table:five}
\end{wraptable}
In addition, to verify the effectiveness of input modalities of RDTTrack, we construct RDTTrack with different input modality combinations. 
The results are shown in Table~\ref{table:three}.
As observed, by leveraging tri-modal input data, RDTTrack outperforms its dual-modal counterparts.
Furthermore, we conduct an experiment to validate the effectiveness of the proposed Depth-TIR module.
The results are exhibited in Table~\ref{table:five}.
As shown, removing the Depth-TIR Orthogonal Projection (OP) leads to a notable performance degradation, with the AUC of RDTTrack dropping significantly from 0.752 to 0.733.
Moreover, when the learnable $\alpha$ and $\beta$ items in Eqn.~\eqref{eq2} are removed, RDTTrack exhibits notable performance drops in both AUC and DP, highlighting their importance in effective feature disentanglement and fusion.

\section{Conclusion}
In this work, we presented the first tri-modal tracking dataset, RGBDT500, and a tri-modal tracker, RDTTrack. 
RGBDT500 is meticulously constructed to address the challenges of tri-modal fusion of RGB, depth, and thermal infrared modalities, offering a comprehensive benchmark for advancing multi-modal tracking research. 
In addition, the proposed RDTTrack effectively leverages the complementary information across all three modalities, achieving state-of-the-art performance on RGBDT500 and validating the benefits of tri-modal integration. 
By releasing this dataset and baseline tracker, we aim to facilitate the development of more robust and generalizable multi-modal tracking methodologies and encourage further exploration into multi-modal fusion for real-world applications.

\begin{ack}
This work was funded by the National Natural Science Foundation of China (62576152, 62020106012, 62336004), the Basic Research Program of Jiangsu (BK20251624, BK20250104), the China Postdoctoral Science Foundation (2025M771593), the Wuxi Science and Technology Development Fund Project (K20241025), and the Fundamental Research Funds for the Central Universities (JUSRP202504007, JUSRP202501041).
\end{ack}

\bibliographystyle{plain}
\bibliography{ref}

@article{xu2021adaptive,
  title={Adaptive channel selection for robust visual object tracking with discriminative correlation filters},
  author={Xu, Tianyang and Feng, Zhenhua and Wu, Xiao-Jun and Kittler, Josef},
  journal={IJCV},
  volume={129},
  number={5},
  pages={1359--1375},
  year={2021},
  publisher={Springer}
}

@article{huang2019got,
  title={Got-10k: A large high-diversity benchmark for generic object tracking in the wild},
  author={Huang, Lianghua and Zhao, Xin and Huang, Kaiqi},
  journal={IEEE TPAMI},
  volume={43},
  number={5},
  pages={1562--1577},
  year={2019},
  publisher={IEEE}
}

@inproceedings{fan2019lasot,
  title={Lasot: A high-quality benchmark for large-scale single object tracking},
  author={Fan, Heng and Lin, Liting and Yang, Fan and Chu, Peng and Deng, Ge and Yu, Sijia and Bai, Hexin and Xu, Yong and Liao, Chunyuan and Ling, Haibin},
  booktitle={CVPR},
  pages={5374--5383},
  year={2019}
}

@inproceedings{yan2021depthtrack,
  title={Depthtrack: Unveiling the power of rgbd tracking},
  author={Yan, Song and Yang, Jinyu and K{\"a}pyl{\"a}, Jani and Zheng, Feng and Leonardis, Ale{\v{s}} and K{\"a}m{\"a}r{\"a}inen, Joni-Kristian},
  booktitle={ICCV},
  pages={10725--10733},
  year={2021}
}

@inproceedings{zhu2023rgbd1k,
  title={RGBD1K: A large-scale dataset and benchmark for RGB-D object tracking},
  author={Zhu, Xue-Feng and Xu, Tianyang and Tang, Zhangyong and Wu, Zucheng and Liu, Haodong and Yang, Xiao and Wu, Xiao-Jun and Kittler, Josef},
  booktitle={AAAA},
  volume={37},
  number={3},
  pages={3870--3878},
  year={2023}
}

@inproceedings{xiao2022attribute,
  title={Attribute-based progressive fusion network for rgbt tracking},
  author={Xiao, Yun and Yang, Mengmeng and Li, Chenglong and Liu, Lei and Tang, Jin},
  booktitle={AAAI},
  volume={36},
  number={3},
  pages={2831--2838},
  year={2022}
}

@article{li2021lasher,
  title={LasHeR: A large-scale high-diversity benchmark for RGBT tracking},
  author={Li, Chenglong and Xue, Wanlin and Jia, Yaqing and Qu, Zhichen and Luo, Bin and Tang, Jin and Sun, Dengdi},
  journal={IEEE TIP},
  volume={31},
  pages={392--404},
  year={2021},
  publisher={IEEE}
}

@article{li2019rgb,
  title={RGB-T object tracking: Benchmark and baseline},
  author={Li, Chenglong and Liang, Xinyan and Lu, Yijuan and Zhao, Nan and Tang, Jin},
  journal={PR},
  volume={96},
  pages={106977},
  year={2019},
  publisher={Elsevier}
}

@article{hartigan1979algorithm,
  title={Algorithm AS 136: A k-means clustering algorithm},
  author={Hartigan, John A and Wong, Manchek A},
  journal={Journal of the Royal Statistical Society},
  volume={28},
  number={1},
  pages={100--108},
  year={1979},
  publisher={JSTOR}
}

@inproceedings{zhu2023visual,
  title={Visual prompt multi-modal tracking},
  author={Zhu, Jiawen and Lai, Simiao and Chen, Xin and Wang, Dong and Lu, Huchuan},
  booktitle={CVPR},
  pages={9516--9526},
  year={2023}
}

@inproceedings{benchmark2016benchmark,
  title={A benchmark and simulator for UAV tracking},
  author={Benchmark, UT},
  booktitle={ECCV},
  year={2016}
}

@inproceedings{lukezic2019cdtb,
  title={Cdtb: A color and depth visual object tracking dataset and benchmark},
  author={Lukezic, Alan and Kart, Ugur and Kapyla, Jani and Durmush, Ahmed and Kamarainen, Joni-Kristian and Matas, Jiri and Kristan, Matej},
  booktitle={ICCV},
  pages={10013--10022},
  year={2019}
}

@inproceedings{song2013tracking,
  title={Tracking revisited using RGBD camera: Unified benchmark and baselines},
  author={Song, Shuran and Xiao, Jianxiong},
  booktitle={ICCV},
  pages={233--240},
  year={2013}
}

@article{xiao2017robust,
  title={Robust fusion of color and depth data for RGB-D target tracking using adaptive range-invariant depth models and spatio-temporal consistency constraints},
  author={Xiao, Jingjing and Stolkin, Rustam and Gao, Yuqing and Leonardis, Ale{\v{s}}},
  journal={IEEE TCYB},
  volume={48},
  number={8},
  pages={2485--2499},
  year={2017},
  publisher={IEEE}
}

@article{li2016learning,
  title={Learning collaborative sparse representation for grayscale-thermal tracking},
  author={Li, Chenglong and Cheng, Hui and Hu, Shiyi and Liu, Xiaobai and Tang, Jin and Lin, Liang},
  journal={IEEE TIP},
  volume={25},
  number={12},
  pages={5743--5756},
  year={2016},
  publisher={IEEE}
}

@inproceedings{qian2021dal,
  title={DAL: A deep depth-aware long-term tracker},
  author={Qian, Yanlin and Yan, Song and Luke{\v{z}}i{\v{c}}, Alan and Kristan, Matej and K{\"a}m{\"a}r{\"a}inen, Joni-Kristian and Matas, Ji{\v{r}}{\'\i}},
  booktitle={ICPR},
  pages={7825--7832},
  year={2021},
}

@inproceedings{danelljan2019atom,
  title={Atom: Accurate tracking by overlap maximization},
  author={Danelljan, Martin and Bhat, Goutam and Khan, Fahad Shahbaz and Felsberg, Michael},
  booktitle={CVPR},
  pages={4660--4669},
  year={2019}
}

@inproceedings{bhat2019learning,
  title={Learning discriminative model prediction for tracking},
  author={Bhat, Goutam and Danelljan, Martin and Gool, Luc Van and Timofte, Radu},
  booktitle={ICCV},
  pages={6182--6191},
  year={2019}
}

@inproceedings{zhao2023arkittrack,
  title={Arkittrack: a new diverse dataset for tracking using mobile RGB-D data},
  author={Zhao, Haojie and Chen, Junsong and Wang, Lijun and Lu, Huchuan},
  booktitle={CVPR},
  pages={5126--5135},
  year={2023}
}

@inproceedings{he2016deep,
  title={Deep residual learning for image recognition},
  author={He, Kaiming and Zhang, Xiangyu and Ren, Shaoqing and Sun, Jian},
  booktitle={CVPR},
  pages={770--778},
  year={2016}
}

@inproceedings{zhang2019multi,
  title={Multi-modal fusion for end-to-end RGB-T tracking},
  author={Zhang, Lichao and Danelljan, Martin and Gonzalez-Garcia, Abel and Van De Weijer, Joost and Shahbaz Khan, Fahad},
  booktitle={ICCVW},
  pages={01--10},
  year={2019}
}

@inproceedings{li2019multi,
  title={Multi-adapter RGBT tracking},
  author={Li, Cheng Long and Lu, Andong and Zheng, Ai Hua and Tu, Zhengzheng and Tang, Jin},
  booktitle={ICCVW},
  pages={2262--2270},
  year={2019},
}

@inproceedings{wu2024single,
  title={Single-model and any-modality for video object tracking},
  author={Wu, Zongwei and Zheng, Jilai and Ren, Xiangxuan and Vasluianu, Florin-Alexandru and Ma, Chao and Paudel, Danda Pani and Van Gool, Luc and Timofte, Radu},
  booktitle={CVPR},
  pages={19156--19166},
  year={2024}
}

@inproceedings{hong2024onetracker,
  title={Onetracker: Unifying visual object tracking with foundation models and efficient tuning},
  author={Hong, Lingyi and Yan, Shilin and Zhang, Renrui and Li, Wanyun and Zhou, Xinyu and Guo, Pinxue and Jiang, Kaixun and Chen, Yiting and Li, Jinglun and Chen, Zhaoyu and others},
  booktitle={CVPR},
  pages={19079--19091},
  year={2024}
}

@inproceedings{wang2024temporal,
  title={Temporal adaptive rgbt tracking with modality prompt},
  author={Wang, Hongyu and Liu, Xiaotao and Li, Yifan and Sun, Meng and Yuan, Dian and Liu, Jing},
  booktitle={AAAI},
  volume={38},
  number={6},
  pages={5436--5444},
  year={2024}
}

@inproceedings{cao2024bi,
  title={Bi-directional adapter for multimodal tracking},
  author={Cao, Bing and Guo, Junliang and Zhu, Pengfei and Hu, Qinghua},
  booktitle={AAAI},
  volume={38},
  number={2},
  pages={927--935},
  year={2024}
}

@inproceedings{hui2023bridging,
  title={Bridging search region interaction with template for rgb-t tracking},
  author={Hui, Tianrui and Xun, Zizheng and Peng, Fengguang and Huang, Junshi and Wei, Xiaoming and Wei, Xiaolin and Dai, Jiao and Han, Jizhong and Liu, Si},
  booktitle={CVPR},
  pages={13630--13639},
  year={2023}
}

@article{dosovitskiy2020image,
  title={An image is worth 16x16 words: Transformers for image recognition at scale},
  author={Dosovitskiy, Alexey},
  journal={arXiv preprint arXiv:2010.11929},
  year={2020}
}

@inproceedings{danelljan2020probabilistic,
  title={Probabilistic regression for visual tracking},
  author={Danelljan, Martin and Gool, Luc Van and Timofte, Radu},
  booktitle={CVPR},
  pages={7183--7192},
  year={2020}
}

@inproceedings{mayer2022transforming,
  title={Transforming model prediction for tracking},
  author={Mayer, Christoph and Danelljan, Martin and Bhat, Goutam and Paul, Matthieu and Paudel, Danda Pani and Yu, Fisher and Van Gool, Luc},
  booktitle={CVPR},
  pages={8731--8740},
  year={2022}
}

@inproceedings{li2017weighted,
  title={Weighted sparse representation regularized graph learning for RGB-T object tracking},
  author={Li, Chenglong and Zhao, Nan and Lu, Yijuan and Zhu, Chengli and Tang, Jin},
  booktitle={ACM MM},
  pages={1856--1864},
  year={2017}
}

@inproceedings{yan2021learning,
  title={Learning spatio-temporal transformer for visual tracking},
  author={Yan, Bin and Peng, Houwen and Fu, Jianlong and Wang, Dong and Lu, Huchuan},
  booktitle={ICCV},
  pages={10448--10457},
  year={2021}
}

@inproceedings{hou2024sdstrack,
  title={Sdstrack: Self-distillation symmetric adapter learning for multi-modal visual object tracking},
  author={Hou, Xiaojun and Xing, Jiazheng and Qian, Yijie and Guo, Yaowei and Xin, Shuo and Chen, Junhao and Tang, Kai and Wang, Mengmeng and Jiang, Zhengkai and Liu, Liang and others},
  booktitle={CVPR},
  pages={26551--26561},
  year={2024}
}

@inproceedings{ye2022joint,
  title={Joint feature learning and relation modeling for tracking: A one-stream framework},
  author={Ye, Botao and Chang, Hong and Ma, Bingpeng and Shan, Shiguang and Chen, Xilin},
  booktitle={ECCV},
  pages={341--357},
  year={2022}
}

@article{lin2017focal,
  title={Focal Loss for Dense Object Detection},
  author={Lin, T},
  journal={arXiv preprint arXiv:1708.02002},
  year={2017}
}

@inproceedings{rezatofighi2019generalized,
  title={Generalized intersection over union: A metric and a loss for bounding box regression},
  author={Rezatofighi, Hamid and Tsoi, Nathan and Gwak, JunYoung and Sadeghian, Amir and Reid, Ian and Savarese, Silvio},
  booktitle={CVPR},
  pages={658--666},
  year={2019}
}

@article{wu2015object,
  title={Object Tracking Benchmark},
  author={Wu, Yi and Lim, Jongwoo and Yang, Ming-Hsuan},
  journal={IEEE TPAMI},
  volume={37},
  number={9},
  pages={1834--1848},
  year={2015},
  publisher={IEEE}
}

@article{xu2023learning,
  title={Learning spatio-temporal discriminative model for affine subspace based visual object tracking},
  author={Xu, Tianyang and Zhu, Xue-Feng and Wu, Xiao-Jun},
  journal={Visual Intelligence},
  volume={1},
  number={1},
  pages={4},
  year={2023},
  publisher={Springer}
}

@article{liu2024spatial,
  title={Spatial-temporal initialization dilemma: towards realistic visual tracking},
  author={Liu, Chang and Yuan, Yongsheng and Chen, Xin and Lu, Huchuan and Wang, Dong},
  journal={Visual Intelligence},
  volume={2},
  number={1},
  pages={35},
  year={2024},
  publisher={Springer}
}

@ARTICLE{javed2023visual,
  author={Javed, Sajid and Danelljan, Martin and Khan, Fahad Shahbaz and Khan, Muhammad Haris and Felsberg, Michael and Matas, Jiri},
  journal={IEEE TPAMI}, 
  title={Visual Object Tracking With Discriminative Filters and Siamese Networks: A Survey and Outlook}, 
  year={2023},
  volume={45},
  number={5},
  pages={6552-6574},
}

@article{liu2024infrared,
  title={Infrared and visible image fusion: From data compatibility to task adaption},
  author={Liu, Jinyuan and Wu, Guanyao and Liu, Zhu and Wang, Di and Jiang, Zhiying and Ma, Long and Zhong, Wei and Fan, Xin},
  journal={IEEE TPAMI},
  year={2024},
  publisher={IEEE}
}

@article{song2025refinefuse,
  title={RefineFuse: an end-to-end network for multi-scale refinement fusion of multi-modality images},
  author={Song, Chengcheng and Li, Hui and Xu, Tianyang and Wu, Xiao-Jun and Kittler, Josef},
  journal={Visual Intelligence},
  volume={3},
  number={1},
  pages={16},
  year={2025},
  publisher={Springer}
}

@inproceedings{zhang2022visible,
  title={Visible-thermal UAV tracking: A large-scale benchmark and new baseline},
  author={Zhang, Pengyu and Zhao, Jie and Wang, Dong and Lu, Huchuan and Ruan, Xiang},
  booktitle={CVPR},
  pages={8886--8895},
  year={2022}
}

@misc{raina2023egoblur,
      title={EgoBlur: Responsible Innovation in Aria},
      author={Nikhil Raina and Guruprasad Somasundaram and Kang Zheng and Sagar Miglani and Steve Saarinen and Jeff Meissner and Mark Schwesinger and Luis Pesqueira and Ishita Prasad and Edward Miller and Prince Gupta and Mingfei Yan and Richard Newcombe and Carl Ren and Omkar M Parkhi},
      year={2023},
      eprint={2308.13093},
      archivePrefix={arXiv},
      primaryClass={cs.CV}
}

@inproceedings{chen2023seqtrack,
  title={Seqtrack: Sequence to sequence learning for visual object tracking},
  author={Chen, Xin and Peng, Houwen and Wang, Dong and Lu, Huchuan and Hu, Han},
  booktitle={CVPR},
  pages={14572--14581},
  year={2023}
}

@article{cui2024mixformer,
  title={MixFormer: End-to-End Tracking With Iterative Mixed Attention},
  author={Cui, Yutao and Jiang, Cheng and Wu, Gangshan and Wang, Limin},
  journal={IEEE TPAMI},
  volume={46},
  number={06},
  pages={4129--4146},
  year={2024},
}

@inproceedings{liu2025dcevo,
  title={DCEvo: Discriminative Cross-Dimensional Evolutionary Learning for Infrared and Visible Image Fusion},
  author={Liu, Jinyuan and Zhang, Bowei and Mei, Qingyun and Li, Xingyuan and Zou, Yang and Jiang, Zhiying and Ma, Long and Liu, Risheng and Fan, Xin},
  booktitle={CVPR},
  pages={2226--2235},
  year={2025}
}

@article{liu2024coconet,
  title={Coconet: Coupled contrastive learning network with multi-level feature ensemble for multi-modality image fusion},
  author={Liu, Jinyuan and Lin, Runjia and Wu, Guanyao and Liu, Risheng and Luo, Zhongxuan and Fan, Xin},
  journal={IJCV},
  volume={132},
  number={5},
  pages={1748--1775},
  year={2024},
  publisher={Springer}
}

@article{liu2024promptfusion,
  title={PromptFusion: Harmonized Semantic Prompt Learning for Infrared and Visible Image Fusion},
  author={Liu, Jinyuan and Li, Xingyuan and Wang, Zirui and Jiang, Zhiying and Zhong, Wei and Fan, Wei and Xu, Bin},
  journal={IEEE/CAA Journal of Automatica Sinica},
  volume={12},
  pages={1--14},
  year={2024},
  publisher={IEEE/CAA Journal of Automatica Sinica}
}

@article{liu2022attention,
  title={Attention-guided global-local adversarial learning for detail-preserving multi-exposure image fusion},
  author={Liu, Jinyuan and Shang, Jingjie and Liu, Risheng and Fan, Xin},
  journal={IEEE TCSVT},
  volume={32},
  number={8},
  pages={5026--5040},
  year={2022},
  publisher={IEEE}
}

@article{jiang2024multispectral,
  title={Multispectral Image Stitching via Global-Aware Quadrature Pyramid Regression},
  author={Jiang, Zhiying and Zhang, Zengxi and Liu, Jinyuan and Fan, Xin and Liu, Risheng},
  journal={IEEE TIP},
 year={2024},
  volume={33},
  number={},
  pages={4288-4302}
}

@article{zhu2022robust,
  author={Zhu, Xue-Feng and Wu, Xiao-Jun and Xu, Tianyang and Feng, Zhen-Hua and Kittler, Josef},
  journal={IEEE TMM}, 
  title={Robust Visual Object Tracking Via Adaptive Attribute-Aware Discriminative Correlation Filters}, 
  year={2022},
  volume={24},
  number={},
  pages={301-312}
}

@article{zhu2024unimod1k,
  title={UniMod1K: Towards a more universal large-scale dataset and benchmark for multi-modal learning},
  author={Zhu, Xue-Feng and Xu, Tianyang and Liu, Zongtao and Tang, Zhangyong and Wu, Xiao-Jun and Kittler, Josef},
  journal={IJCV},
  volume={132},
  number={8},
  pages={2845--2860},
  year={2024},
  publisher={Springer}
}

@article{zhu2024self,
  title={Self-supervised learning for RGB-D object tracking},
  author={Zhu, Xue-Feng and Xu, Tianyang and Atito, Sara and Awais, Muhammad and Wu, Xiao-Jun and Feng, Zhenhua and Kittler, Josef},
  journal={PR},
  volume={155},
  pages={110543},
  year={2024},
  publisher={Elsevier}
}

@inproceedings{tang2024generative,
  title={Generative-based fusion mechanism for multi-modal tracking},
  author={Tang, Zhangyong and Xu, Tianyang and Wu, Xiaojun and Zhu, Xue-Feng and Kittler, Josef},
  booktitle={AAAI},
  volume={38},
  number={6},
  pages={5189--5197},
  year={2024}
}

@article{tang2025revisiting,
  title={Revisiting rgbt tracking benchmarks from the perspective of modality validity: A new benchmark, problem, and solution},
  author={Tang, Zhangyong and Xu, Tianyang and Wu, Xiao-Jun and Zhu, Xue-Feng and Cheng, Chunyang and Feng, Zhenhua and Kittler, Josef},
  journal={IEEE TIP},
  year={2025}
}

@article{tang2025omni,
  title={Omni Survey for Multimodality Analysis in Visual Object Tracking},
  author={Tang, Zhangyong and Xu, Tianyang and Zhu, Xue-Feng and Li, Hui and Zhao, Shaochuan and Zhou, Tao and Cheng, Chunyang and Wu, Xiao-Jun and Kittler, Josef},
  journal={arXiv preprint arXiv:2508.13000},
  year={2025}
}

@inproceedings{cheng2025one,
  title={One model for all: Low-level task interaction is a key to task-agnostic image fusion},
  author={Cheng, Chunyang and Xu, Tianyang and Feng, Zhenhua and Wu, Xiaojun and Tang, Zhangyong and Li, Hui and Zhang, Zeyang and Atito, Sara and Awais, Muhammad and Kittler, Josef},
  booktitle={Proceedings of the IEEE/CVF Conference on Computer Vision and Pattern Recognition},
  pages={28102--28112},
  year={2025}
}

@inproceedings{kristan2023first,
  title={The first visual object tracking segmentation vots2023 challenge results},
  author={Kristan, Matej and Matas, Ji{\v{r}}{\'\i} and Danelljan, Martin and Felsberg, Michael and Chang, Hyung Jin and Zajc, Luka {\v{C}}ehovin and Luke{\v{z}}i{\v{c}}, Alan and Drbohlav, Ondrej and Zhang, Zhongqun and Tran, Khanh-Tung and others},
  booktitle={Proceedings of the IEEE/CVF international conference on computer vision},
  pages={1796--1818},
  year={2023}
}

@article{xu2020accelerated,
  title={An accelerated correlation filter tracker},
  author={Xu, Tianyang and Feng, Zhen-Hua and Wu, Xiao-Jun and Kittler, Josef},
  journal={Pattern recognition},
  volume={102},
  pages={107172},
  year={2020},
  publisher={Elsevier}
}

%%%%%%%%%%%%%%%%%%%%%%%%%%%%%%%%%%%%%%%%%%%%%%%%%%%%%%%%%%%%
\newpage
\appendix

\section{Technical Appendices and Supplementary Material}
\subsection{Data Samples}

\begin{figure*}[h]
  \centering
   \includegraphics[width=1.0\linewidth]{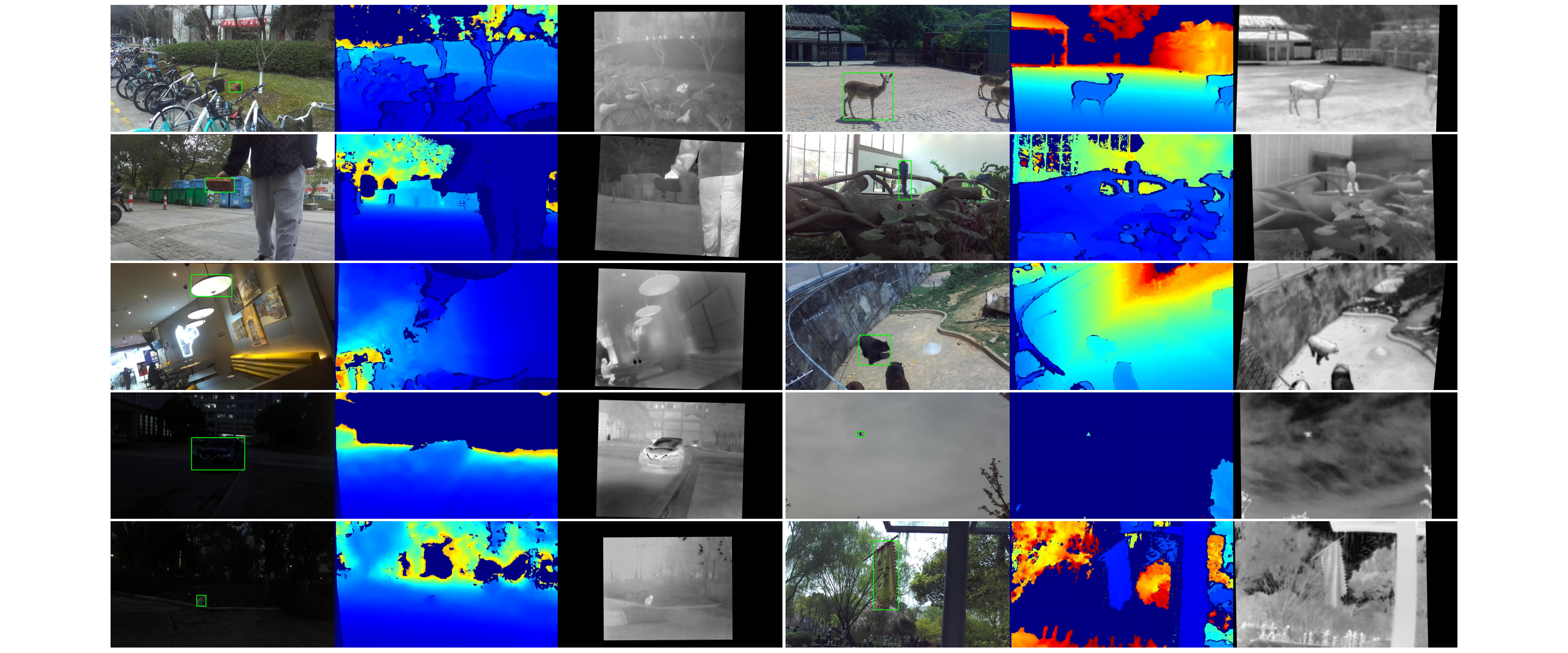}
   \caption{Some samples from the RGBDT500 dataset. The targeted object are highlighted by green bounding boxes.}
   \label{fig:samples}
\end{figure*}

Some samples from the RGBDT500 dataset are presented in Fig.~\ref{fig:samples}.
As shown, the proposed RGBDT500 covers a wide range of scenarios and includes diverse object categories, highlighting its versatility for multi-modal tracking task.
In these scenarios, the three modalities provide complementary information, which can contribute to the improvement of tracking performance.

\subsection{Qualitative Analysis}

\begin{figure}[h]
  \centering
   \includegraphics[trim={0mm 118mm 0mm 0mm}, width=1.0\linewidth]{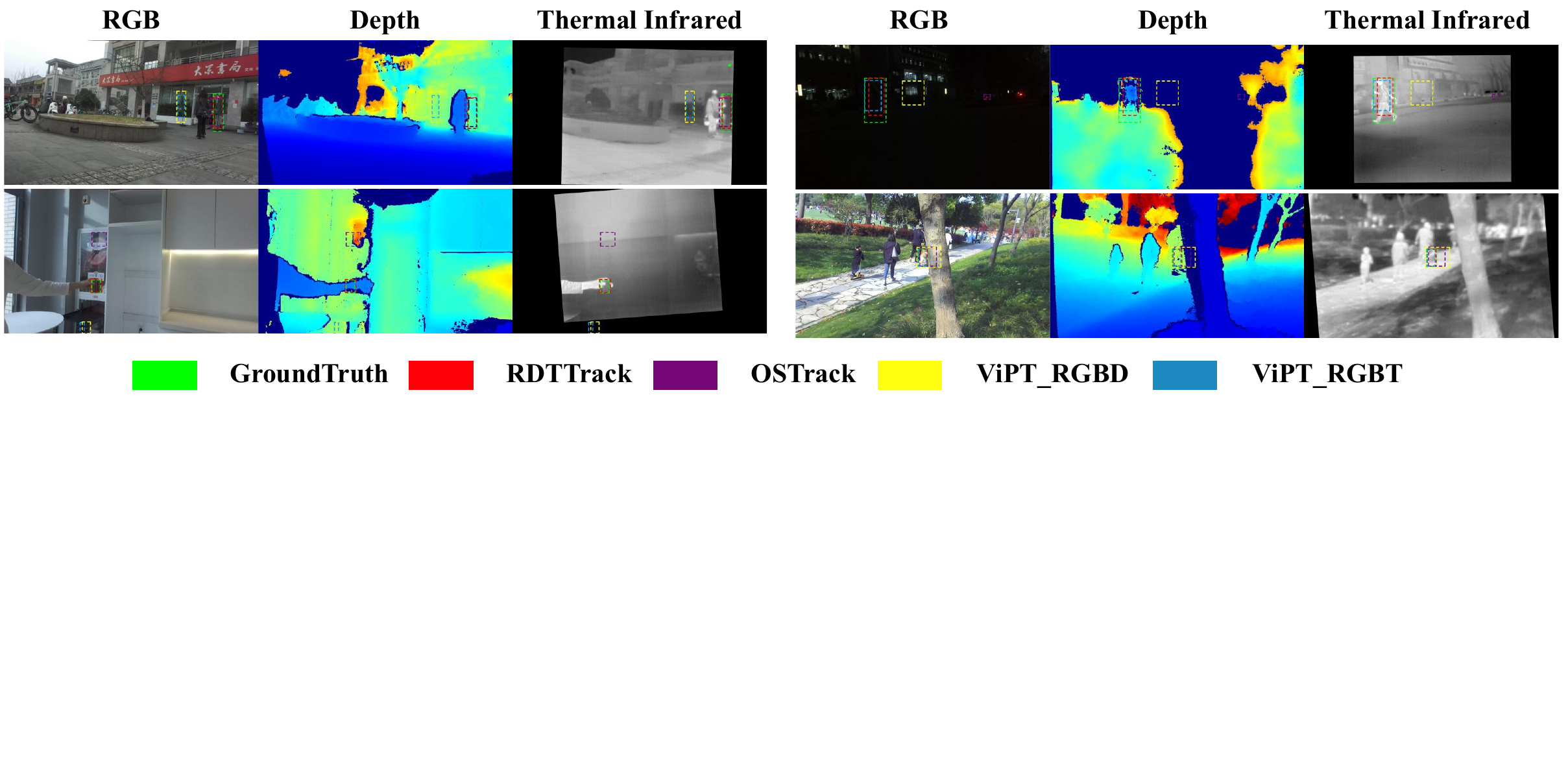}
   \caption{Visualisation of the results of several mainstream trackers on sequences of RGBDT500.}
   \label{fig:vis}
\end{figure}

In Fig.~\ref{fig:vis}, we visualized the tracking results of some trackers, including OsTrack (RGB)~\citep{ye2022joint}, ViPT\_RGBD (RGB + Depth), ViPT\_RGBT (RGB + Thermal)~\citep{zhu2023visual}, and RDTTrack (RGB + Depth + Thermal) on several complex frames of RGBDT500. 
As shown, in challenging scenarios such as low-light environments or modality interference, our RDTTrack effectively leverages the most reliable modality in each scene to achieve more robust and accurate tracking.

\subsection{Limitations and Future Work}
Despite the contributions presented in this work, we acknowledge several limitations that pave the way for future research.
Specifically, the proposed RDTTrack baseline lacks the flexibility to handle a variable number of input modalities. 
This restricts its deployment in real-world applications where sensor configurations may vary. 
In addition, the RGBDT500 dataset, while a significant step forward, could be extended to incorporate other vital modalities, such as LiDAR or event. 
Future efforts will be directed towards designing a more flexible and input-agnostic fusion architecture, as well as expanding the dataset to include a wider array of sensing modalities.

\subsection{Privacy Preservation}
The RGBDT500 dataset was collected in controlled environments, with a focus on non-identifiable household objects and animals. 
In rare scenarios where individuals appear in the dataset, those individuals were members of our research and data collection team, who were fully aware of the recording and gave explicit consent to be included. 
No bystanders or members of the public were captured without consent. 
Furthermore, all frames were reviewed to ensure that no identifiable facial features or license plates are present in the released dataset. 
The potential identifiable facial and license plates are anonymized through blurring by using EgoBlur model. 

In addition, all sequences involving public capture occurred in areas where photography is legally permitted. For any such sequences, signage was displayed to indicate that recording was taking place for academic research purposes. 
Additionally, we have set up a dedicated contact email (xuefeng.zhu@jiangnan.edu.cn) on the dataset website, where individuals can request takedown of specific sequences if they believe they are represented. 

\subsection{Broader Impacts and Safeguards} 

This work enhances the robustness of multi-modal tracking, presenting both potential positive and negative societal impacts. 
On the positive side, it can enable life-saving applications such as improving autonomous vehicle safety in adverse weather conditions and advancing wildlife monitoring efforts. 
Conversely, the enhanced tracking capabilities across RGB, depth, and thermal domains may be misused in surveillance systems, including mass monitoring, protest tracking, or law enforcement applications that could infringe on privacy and civil liberties. 
Due to the dual-use nature of this technology, it requires responsible development frameworks and ethical deployment guidelines to mitigate risks such as privacy violations.
To mitigate such risks, the RGBDT500 dataset is released under a research-only, non-commercial license that explicitly prohibits use for surveillance, military, or law enforcement purposes. 
Access to the download link of RGBDT500 requires click-through acceptance of these terms. 
All sequences involving individuals feature team members with explicit consent, and no identifiable information is present. We encourage transparency and welcome community feedback to support responsible use.

\subsection{Dataset Card}
\begin{itemize}
\item  Geographic distribution: Data was collected across distinct regions, including urban, suburban, and indoor lab settings, representing diverse environments and sensor conditions.
\item  Day/Night ratio: Approximately 80\% of the sequences were recorded during the day and 20\% at night, to reflect a variety of lighting scenarios.
\item  Demographic information: As the dataset primarily includes objects and animals in controlled environments, demographic attributes are not labelled or applicable. In rare cases involving people, subjects were team members or individuals who provided consent, and no demographic classification was performed.
\item  Fairness audit plan: While demographic analysis is not applicable for this version of the dataset, we include a short plan to expand fairness auditing in future iterations. This includes evaluating the dataset’s impact in downstream models, e.g., bias in object tracking under varying conditions, and inviting feedback from users to inform gaps in representativeness.
\end{itemize}

\end{document}